# Fighting an Infodemic: COVID-19 Fake News Dataset

**Parth Patwa[§], Shivam Sharma*,[†], Srinivas PYKL[§], Vineeth Guptha*, Gitanjali Kumari[‡], Md Shad Akhtar[†], Asif Ekbal[‡], Amitava Das*, Tanmoy Chakraborty[†]**

[§]*IIIT Sri City*, India. [†]*IIIT Delhi*, India [‡]*IIT Patna*, India. [*]*Wipro Reseach*, India
{parthprasad.p17, srinivas.p}@iiits.in
{tanmoy, shad.akhtar}@iiitd.ac.in
{gitanjali_2021cs03, asif}@iitp.ac.in
{shivam.sharma23, bodla.guptha, amitava.das2}@wipro.com

## Abstract

Along with COVID-19 pandemic we are also fighting an 'infodemic'. Fake news and rumors are rampant on social media. Believing in rumors can cause significant harm. This is further exacerbated at the time of a pandemic. To tackle this, we curate and release a manually annotated dataset of 10,700 social media posts and articles of real and fake news on COVID-19. We perform a binary classification task (real vs fake) and benchmark the annotated dataset with four machine learning baselines - Decision Tree, Logistic Regression, Gradient Boost, and Support Vector Machine (SVM). We obtain the best performance of 93.32% F1-score with SVM on the test set. The data and code is available at: https://github.com/parthpatwa/covid19-fake-news-dectection. [1]

## 1 Introduction

The use of social media is increasing with the time. In 2020, there are over 3.6 billion users on social media, and by 2025, it is expected that there will be around 4.41 billion users (J.Cement, 2020). Social media has brought us many benefits like, faster and easier communication, brand promotions, customer feedback, etc; however, it also has several disadvantages, and one of the prominent ones being fake news. Fake news is unarguably a threat to the society (Panke, 2020) and it has become a challenging problem for social media users and researchers alike. Fake news on COVID-19 is a much bigger problem as it can influence people to take extreme measures by believing that the news is true. For example, a fake news ('*Alcohol is a cure for COVID-19*') led to many deaths and hospitalizations in Iran (Karimi and Gambrell, 2020). This shows how vulnerable we are to fake-news in these hard times and how severe the outcome can be, if we ignore them. The first step towards tackling the fake news is to identify it. We primarily restrict our investigation of the social media content to the topic COVID-19.

To tackle fake news, in this paper, we present a dataset of social media posts and articles on COVID-19 with real and fake labels. The targeted media platforms for the data collection are designated to be the ones that are actively used for social networking for peer communication and relaying information, which could be in the form of news, events, social phenomenon, etc. We collect both real news and fake claims that surfaced on social media on COVID-19 topic. Fake claims are collected from various fact-checking websites like Politifact[2], NewsChecker[3], Boomlive[4], etc., and from tools like Google fact-check-explorer[5] and IFCN chatbot[6]. Real news is collected from Twitter using verified twitter handles. We also perform exploratory data analysis and implement four machine learning baselines.

## 2 Related Work

There is no agreement amongst researchers over the definition of *fake news*. A simple definition of fake news is the news which is intentionally created false as news articles to mislead readers. It is adopted in various recent research (Mustafaraj and Metaxas, 2017; Potthast et al., 2018). In another definition, deceptive news which includes news fabrications, satire, hoaxes, etc., are considered as fake news (Balmas, 2014; Rubin et al., 2016). Despite the existence of several works dedicated for fake news, accurate automatic fake news detection is an extremely challenging task. The lack of a common acceptable benchmark dataset for this task is one of the key problems.

[2]http://www.politifact.com
[3]https://newschecker.in/
[4]www.boomlive.in
[5]https://toolbox.google.com/factcheck/explorer
[6]http://poy.nu/ifcnbot

| Label | Source | Text |
|---|---|---|
| Fake | Facebook | All hotels, restaurants, pubs etc will be closed till 15th Oct 2020 as per tourism minister of India. |
| Fake | Twitter | #Watch Italian Billionaire commits suicide by throwing himself from 20th Floor of his tower after his entire family was wiped out by #Coronavirus #Suicide has never been the way, may soul rest in peace May God deliver us all from this time |
| Fake | Fact checking | Scene from TV series viral as dead doctors in Italy due to COVID-19 |
| Fake | Twitter | It's being reported that NC DHHS is telling hospitals that if they decide to do elective surgeries, they won't be eligible to receive PPE from the state. The heavy hand of government. I hope Secretary Cohen will reverse course. #NCDHHS #COVID19NC #ncpol |
| Real | Twitter (WHO) | Almost 200 vaccines for #COVID19 are currently in clinical and pre-clinical testing. The history of vaccine development tells us that some will fail and some will succeed-@DrTedros #UNGA #UN75 |
| Real | Twitter (CDC) | Heart conditions like myocarditis are associated with some cases of #COVID19. Severe cardiac damage is rare but has occurred even in young healthy people. CDC is working to understand how COVID-19 affects the heart and other organs. |
| Real | Twitter (ICMR) | ICMR has approved 1000 #COVID19 testing labs all across India. There was only one government lab at the beginning of the year. #IndiaFightsCorona. #ICMRFightsCovid19 |

Table 1: Examples of real and fake news from the dataset. Fake news is collected from various sources. Real news is collected from verified twitter accounts.

In 2017, Pomerleau and Rao organized the Fake News Challenge (FNC-1)[7] task to address the issue. Authors proposed a new dataset using which both in-domain and cross-domain experiments are done with the help of machine learning and deep learning techniques to automatically detect fake news. (Vlachos and Riedel, 2014) collected 221 labeled claims that were fact checked by journalists available online. For this purpose, the fact checking blog of Channel45 and the Truth-O-Meter from PolitiFact is used. (Zubiaga et al., 2016) presented a methodology to identify and annotate a dataset having 330 rumour threads (4,842 tweets) associated with 9 newsworthy events. The intuition is to understand how social media users spread, support, or deny rumours before and after its veracity status is resolved. In a similar way, several approaches have been developed to identify as-well-as limit the spread of (mis-)information (Acemoglu et al., 2010; Budak et al., 2011; Kwon et al., 2013; Ma et al., 2017; Nguyen et al., 2020; Chandra et al., 2020). Vo and Lee 2020 (Vo and Lee, 2020) approach the problem of fake news spread prevention by proposing a multimodal attention network that learns to rank the fact-checking documents based on their relevancy.

For rumour debunking, (Ferreira and Vlachos, 2016) created a dataset named *Emergent*, which is derived from a digital journalism project. The dataset contains 300 rumoured claims and 2,595 associated news articles. This dataset is collected and annotated by journalists with an estimation of their veracity (true, false, or unverified) similar to fake news detection task. Researchers have also shown their interest in automatic detection of deceptive content for the domains such as consumer review websites, online advertising, online dating, etc. (Zhang and Guan, 2008; Warkentin et al., 2010; Shafqat et al., 2016).

In addition to the above works, scientists have also been trying to discover AI related techniques to deal with 'infodemic' of misinformation related to the COVID-19 pandemic. (Shahi and Nandini, 2020) presented a multilingual cross-domain dataset of 5182 fact-checked news articles for COVID-19. They collected the articles from 92 different fact-checking website. (Kar et al., 2020) proposed a BERT based model augmented with additional features extracted from Twitter to identify fake tweets related to COVID-19. They also used mBERT model for multiple Indic Language. (Vijjali et al., 2020) developed an automated

[7]http://www.fakenewschallenge.org

pipeline for COVID-19 fake news detection using fact checking algorithms and textual entailment.

## 3 Dataset Development

We curate a dataset of real and fake news on COVID-19:

- **Real** - Tweets from verified sources and give useful information on COVID-19.
- **Fake** - Tweets, posts, articles which make claims and speculations about COVID-19 which are verified to be not true.

Table 1 gives some examples of real and fake news from the dataset.

### 3.1 Collection and Annotation

We follow a simple guideline during the data collection phase as follows:

- Content is related to the topic of COVID-19.
- Only textual English contents are considered. Non-English posts are skipped. Language is detected manually.

### 3.2 Fake News

We collect fake news data from public fact-verification websites and social media. The posts are manually verified with the original documents. Various web based resources like Facebook posts, tweets, a news piece, Instragram posts, public statements, press releases, or any other popular media content, are leveraged towards collecting fake news content. Besides these, popular fact-verification websites like PolitiFact, Snopes[8], Boomlive are also used as they play a crucial role towards collating the manually adjudicated details of the veracity of the claims becoming viral. These websites host COVID-19 and other generic topic related verdicts. The factually verified (fake) content can be easily found from such websites.

### 3.3 Real News

To collect potential real tweets, we first crawl tweets from official and verified twitter handles of the relevant sources using twitter API. The relevant sources are the official government accounts, medical institutes, news channels, etc. We collect tweets from 14 such sources, e.g., World Health Organization (WHO), Centers for Disease Control and Prevention (CDC), Covid India Seva, Indian Council of Medical Research (ICMR), etc. Each tweets is read by a human and is marked as real news if it contains useful information on COVID-19 such as numbers, dates, vaccine progress, government policies, hotspots, etc.

[8]http://www.snopes.com/

| Attribute | Fake | Real | Combined |
|---|---|---|---|
| Unique words | 19728 | 22916 | 37503 |
| Avg words per post | 21.65 | 31.97 | 27.05 |
| Avg chars per post | 143.26 | 218.37 | 182.57 |

Table 2: Numeric features of the dataset

### 3.4 Dataset Statistics

From Table 2, we observe that, in general, real news are longer than fake news in terms of average number of words and characters per post. The vocabulary size (i.e., unique words) of the dataset is 37,505 with 5141 common words in both fake and real news.

The dataset is split into train (60%), validation (20%), test (20%). Table 3 shows the class-wise distribution of all data splits. The dataset is class-wise balanced as 52.34% of the samples consist of real news and 47.66% of the data consists of fake news. Moreover, we maintain the class-wise distribution across train, validation, and test splits.

We also analyse the dataset on token-level. The 10 most frequent tokens after removing stopwords are:

- Fake: coronavirus, covid19, people, will, new, trump, says, video, vaccine, virus.
- Real: covid19, cases, new, tests, number, total, people, reported, confirmed, states.
- Combined: covid19, cases, coronavirus, new, people, tests, number, will, deaths, total.

Figures 1a, 1b and 1c show word clouds for fake news, real news, and combined data respectively. From the word clouds and most frequent words, we see that there is a significant overlap of important words across fake and real news.

## 4 Baselines and Results

**Pre-processing** - We remove all the links, non alphanumeric characters and English stop words.

**Feature extraction**- We use term fre- quency–inverse document frequency (tf-idf) for feature extraction. tf-idf for a word increases with its frequency in a document and decreases as the number of documents in the corpus that contain the word increases.

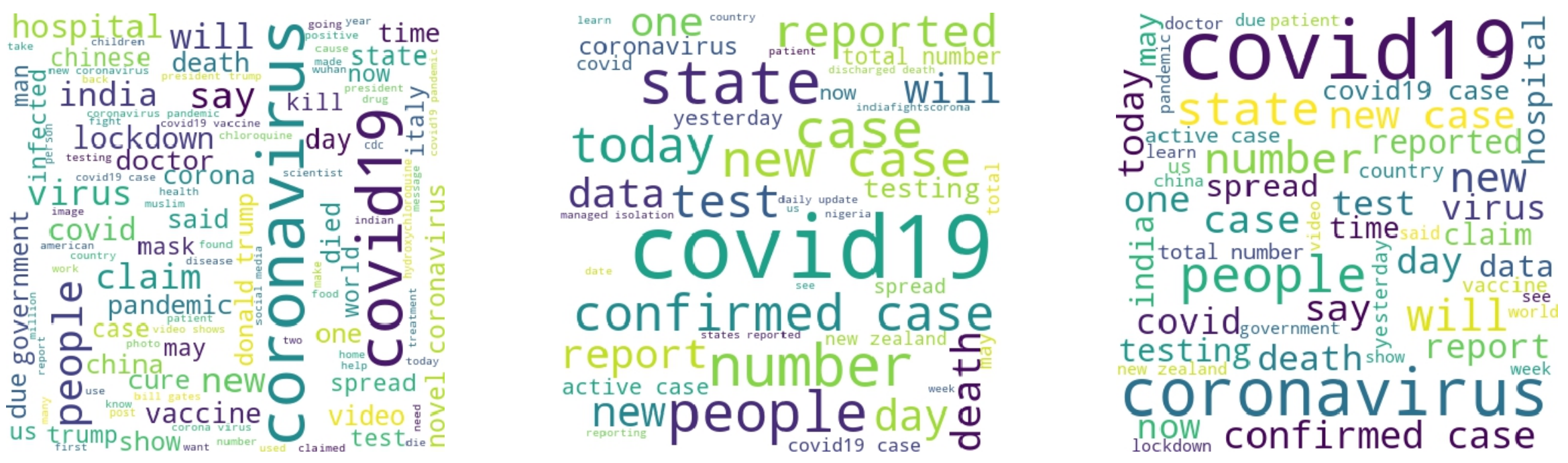

(a) Fake news

(b) Real news

(c) Fake+real news

Figure 1: Word clouds generated from the dataset.

| Split | Real | Fake | Total |
|---|---|---|---|
| Training | 3360 | 3060 | 6420 |
| Validation | 1120 | 1020 | 2140 |
| Test | 1120 | 1020 | 2140 |
| **Total** | 5600 | 5100 | 10700 |

Table 3: distribution of data across classes and splits. Note that the data is class-wise balanced and the class-wise distribution is similar across splits.

| Model | Acc | P | R | F1 |
|---|---|---|---|---|
| DT | 85.23 | 85.31 | 85.23 | 85.25 |
| LR | 92.76 | 92.79 | 92.76 | 92.75 |
| **SVM** | **93.46** | **93.48** | **93.46** | **93.46** |
| GDBT | 86.82 | 87.08 | 86.82 | 86.82 |

Table 4: Accuracy (Acc), weighted average Precision (P), weighted average Recall (P) and weighted average f1 score (f1) of ML models on the *validation* data.

| Model | Acc | P | R | F1 |
|---|---|---|---|---|
| DT | 85.37 | 85.47 | 85.37 | 85.39 |
| LR | 91.96 | 92.01 | 91.96 | 91.96 |
| **SVM** | **93.32** | **93.33** | **93.32** | **93.32** |
| GDBT | 86.96 | 87.24 | 86.96 | 86.96 |

Table 5: Accuracy (Acc), weighted average Precision (P), weighted average Recall (P) and weighted average f1 score (f1) of ML models on the *test* data.

**ML Algorithms** - for classification, we experiment with Logistic Regression (LR), Support Vector Machine (SVM) with linear kernel, Decision Tree (DT) and Gradient Boost (GDBT). All algorithms are implemented using sklearn package. All experiments run in approx 1 minute on an i7 CPU. The code is available on github.[9]

Table 4 shows the results of ML models on validation dataset whereas table 5 shows the results on test dataset. From table 5 we can see that the best test F1 score of 93.32% is achieved by SVM closely followed by Logistic Regression (LR) with 91.96% F1-score. In comparison, Decision Tree (DT) and Gradient Boost (GDBT) reported significantly inferior performance with 85.39% and 86.96% F1-scores, respectively. The results are similar for validation set, which shows that the distributions of test set and validation set are similar. For all the models, the respective precision and recall are close to each other.

Figures 2a and 2b show the confusion matrix of the predictions of SVM on the validation and test set respectively. Since we train, validate, and test the model on a balanced dataset, the predictions are also balanced across two labels.

## 5 Conclusion and Future Work

In this paper, we describe and release a fake news detection dataset containing 10,700 fake and real news related to COVID-19. We collect these posts from various social media and fact checking websites, and manually verify the veracity of each posts. The data is class-wise balanced and can be used to develop automatic fake news and rumor detection algorithms. We also benchmark the developed dataset using machine learning algorithm and project them as the potential baselines. Among the machine learning models, SVM-based classifier performs the best with 93.32% F1-score on the test

[9]https://github.com/parthpatwa/covid19-fake-news-detection

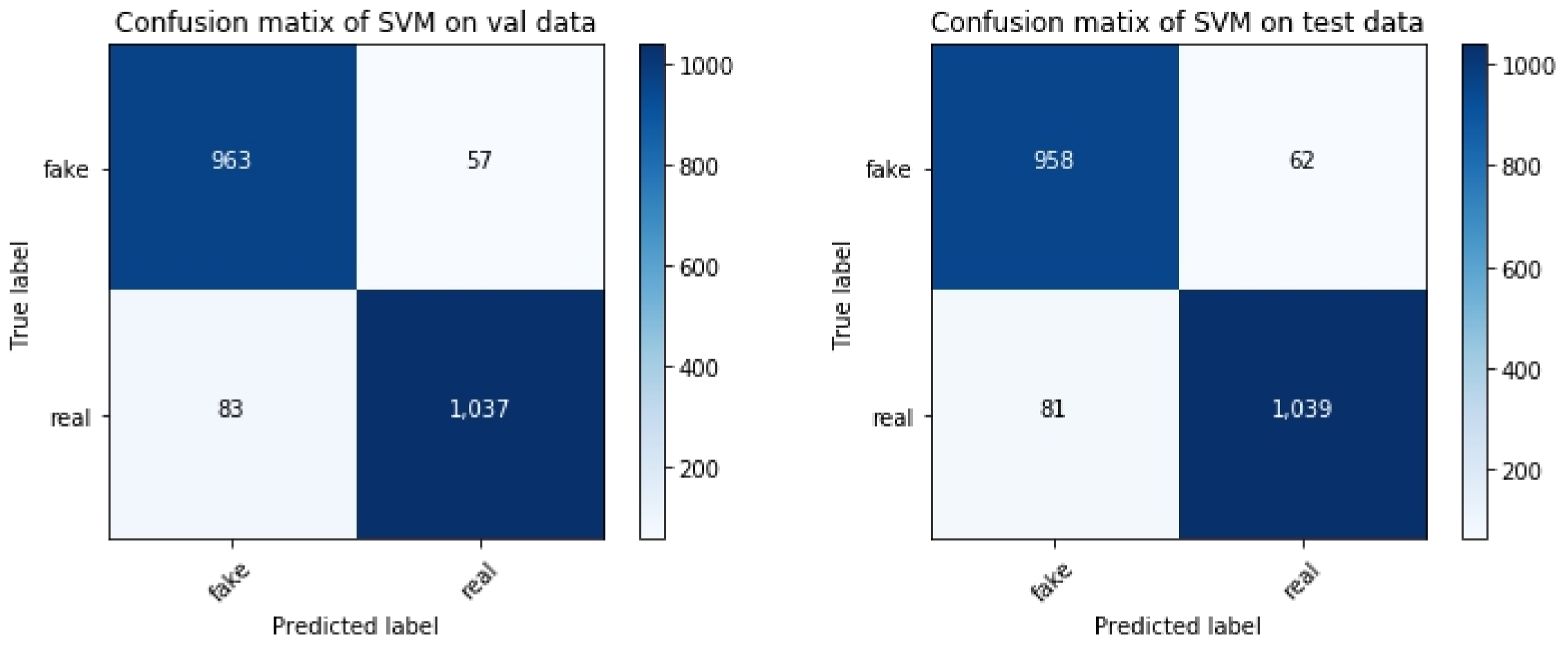

(a) Validation data (b) Test data

Figure 2: Confusion matrix of SVM model on the validation and test datasets respectively. Performance on both the classes is similar.

set.

Future work could be targeted towards collecting more data, enriching the data by providing the reason for being real/fake along with the labels, collecting multilingual data. Using deep learning instead of machine learning is also worth exploring.